\documentclass[sigconf]{acmart}

\usepackage{graphicx}
\usepackage{tabularx}
\usepackage[mathscr]{euscript}
\usepackage{amsmath}
\usepackage{multirow}
\usepackage{amsfonts}
\usepackage{float}

\newcolumntype{x}[1]{>{\centering\arraybackslash}p{#1}}
\newcolumntype{Y}{>{\centering\arraybackslash}X}

\usepackage{balance}
\def\BibTeX{{\rm B\kern-.05em{\sc i\kern-.025em b}\kern-.08emT\kern-.1667em\lower.7ex\hbox{E}\kern-.125emX}}

\begin{document}

%


\copyrightyear{2019} 
\acmYear{2019} 
\setcopyright{acmcopyright}
\acmConference[KDD '19]{The 25th ACM SIGKDD Conference on Knowledge Discovery and Data Mining}{August 4--8, 2019}{Anchorage, AK, USA}
\acmBooktitle{The 25th ACM SIGKDD Conference on Knowledge Discovery and Data Mining (KDD '19), August 4--8, 2019, Anchorage, AK, USA}
\acmPrice{15.00}
\acmDOI{10.1145/3292500.3330897}
\acmISBN{978-1-4503-6201-6/19/08}

\settopmatter{printacmref=true}
\fancyhead{}

%
\title{Graph Representation Learning via Hard and Channel-Wise Attention Networks}

%

\author{Hongyang Gao}
\affiliation{%
  \institution{Texas A\&M University}
  \city{College Station}
  \state{TX}
  \postcode{77843}
}
\email{hongyang.gao@tamu.edu}

\author{Shuiwang Ji}
\affiliation{%
  \institution{Texas A\&M University}
  \city{College Station}
  \state{TX}
  \postcode{77843}
}
\email{sji@tamu.edu}

%
\renewcommand{\shortauthors}{H. Gao, et al.}

\begin{abstract}

Attention operators have been widely applied in various fields,
including computer vision, natural language processing, and network
embedding learning. Attention operators on graph data enables
learnable weights when aggregating information from neighboring
nodes. However, graph attention operators (GAOs) consume excessive
computational resources, preventing their applications on large
graphs. In addition, GAOs belong to the family of soft attention,
instead of hard attention, which has been shown to yield better
performance. In this work, we propose novel hard graph attention
operator~(hGAO) and channel-wise graph attention operator~(cGAO).
hGAO uses the hard attention mechanism by attending to only
important nodes. Compared to GAO, hGAO improves performance and
saves computational cost by only attending to important nodes. To
further reduce the requirements on computational resources, we
propose the cGAO that performs attention operations along channels.
cGAO avoids the dependency on the adjacency matrix, leading to
dramatic reductions in computational resource requirements.
Experimental results demonstrate that our proposed deep models with
the new operators achieve consistently better performance.
Comparison results also indicates that hGAO achieves significantly
better performance than GAO on both node and graph embedding tasks.
Efficiency comparison shows that our cGAO leads to dramatic savings
in computational resources, making them applicable to large graphs.

\end{abstract}

%
%



\maketitle


\section{Introduction}

Deep attention networks are becoming increasingly powerful in
solving challenging tasks in various fields, including natural
language
processing~\cite{vaswani2017attention,luong2015effective,devlin2018bert},
and computer vision~\cite{wang2018non,xu2015show,zhao2018psanet}.
Compared to convolution layers and recurrent neural layers like
LSTM~\cite{hochreiter1997long,gregor2015draw}, attention operators
are able to capture long-range dependencies and relationships among
input elements, thereby boosting
performance~\cite{devlin2018bert,li2018non}. In addition to images
and texts, attention operators are also applied on
graphs~\cite{velivckovic2017graph}. In graph attention
operators~(GAOs), each node in a graph attend to all neighboring
nodes, including itself. By employing attention mechanism, GAOs
enable learnable weights for neighboring feature vectors when
aggregating information from neighbors. However, a practical
challenge of using GAOs on graph data is that they consume excessive
computational resources, including computational cost and memory
usage. The time and space complexities of GAOs are both quadratic to
the number of nodes in graphs. At the same time, GAOs belong to the
family of soft attention~\cite{jaderberg2015spatial}, instead of
hard attention~\cite{xu2015show}. It has been shown that hard
attention usually achieves better performance than soft attention,
since hard attention only attends to important
features~\cite{shankar2018surprisingly,xu2015show,yu2019st}.

In this work, we propose novel hard graph attention operator~(hGAO).
hGAO performs attention operation by requiring each query node to
only attend to part of neighboring nodes in graphs. By employing a
trainable project vector $\boldsymbol p$, we compute a scalar
projection value of each node in graph on $\boldsymbol p$. Based on
these projection values, hGAO selects several important neighboring
nodes to which the query node attends. By attending to the most
important nodes, the responses of the query node are more accurate,
thereby leading to better performance than methods based on soft
attention. Compared to GAO, hGAO also saves computational cost by
reducing the number of nodes to attend.

GAO also suffers from the limitations of excessive requirements on
computational resources, including computational cost and memory
usage. hGAO improves the performance of attention operator by using
hard attention mechanism. It still consumes large amount of memory,
which is critical when learning from large graphs. To overcome this
limitation, we propose a novel channel-wise graph attention
operator~(cGAO). cGAO performs attention operation from the
perspective of channels. The response of each channel is computed by
attending to all channels. Given that the number of channels is far
smaller than the number of nodes, cGAO can significantly save
computational resources. Another advantage of cGAO over GAO and hGAO
is that it does not rely on the adjacency matrix. In both GAO and
hGAO, the adjacency matrix is used to identify neighboring nodes for
attention operators. In cGAO, features within the same node
communicate with each other, but features in different nodes do not.
cHAO does not need the adjacency matrix to identify nodes
connectivity. By avoiding dependency on the adjacency matrix, cGAO
achieves better computational efficiency than GAO and hGAO.

Based on our proposed hGAO and cGAO, we develop deep attention
networks for graph embedding learning. Experimental results on graph
classification and node classification tasks demonstrate that our
proposed deep models with the new operators achieve consistently
better performance. Comparison results also indicates that hGAO
achieves significantly better performance than GAOs on both node and
graph embedding tasks. Efficiency comparison shows that our cGAO
leads to dramatic savings in computational resources, making them
applicable to large graphs.

\section{Background and Related Work}\label{sec:relatedwork}

In this section, we describe the attention operator and related hard
attention and graph attention operators.

\subsection{Attention Operator}\label{sec:att}

An attention operator takes three matrices as input; those are a query matrix
$\boldsymbol Q = [\mathbf{q}_{1}, \mathbf{q}_{2}, \cdots, \mathbf{q}_{m}] \in
\mathbb{R}^{d \times m}$ with each $\mathbf{q}_{i} \in \mathbb{R}^{d}$, a key
matrix $\boldsymbol K = [\mathbf{k}_{1}, \mathbf{k}_{2}, \cdots,
\mathbf{k}_{n}] \in \mathbb{R}^{d \times n}$ with each $\mathbf{k}_{i} \in
\mathbb{R}^{d}$, and a value matrix $\boldsymbol V = [\mathbf{v}_{1},
\mathbf{v}_{2}, \cdots, \mathbf{v}_{n}] \in \mathbb{R}^{p \times n}$ with each
$\mathbf{v}_{i} \in \mathbb{R}^{p}$. For each query vector $\boldsymbol q_i$,
the attention operator produces its response by attending it to every key vector
in $\boldsymbol K$. The results are used to compute a weighted sum of all value
vectors in $\boldsymbol V$, leading to the output of the attention operator.
The layer-wise forward-propagation operation of attn($\boldsymbol Q$, $\boldsymbol K$,
$\boldsymbol V$) is defined as
\begin{equation}\label{eq:att}
\begin{aligned}
\boldsymbol E &= \boldsymbol K^T \boldsymbol Q &\in& \mathbb{R}^{n \times m},\\
\boldsymbol O &= \boldsymbol V \mbox{softmax}(\boldsymbol E) &\in& \mathbb{R}^{p \times m},
\end{aligned}
\end{equation}
where $\mbox{softmax}(\cdot)$ is a column-wise softmax operator.

The coefficient matrix $\boldsymbol E$ is calculated by the matrix
multiplication between $\boldsymbol K^T$ and $\boldsymbol Q$. Each
element $e_{ij}$ in $\boldsymbol E$ represents the inner product
result between the key vector $\boldsymbol k^T_i$ and the query
vector $\boldsymbol q_j$. The matrix multiplication $\boldsymbol K^T
\boldsymbol Q$ computes all similarity scores between all query
vectors and all key vectors. The column-wise softmax operator is
used to normalize the coefficient matrix and make the sum of each
column to 1. The matrix multiplication between $\boldsymbol V$ and
$\mbox{softmax}(\boldsymbol E)$ produces the output $\boldsymbol O$.
Self-attention ~\cite{vaswani2017attention} is a special attention
operator with $\boldsymbol Q=\boldsymbol K=\boldsymbol V$.

In Eq.~\ref{eq:att}, we employ dot product to calculate responses
between query vectors in $\boldsymbol Q$ and key vectors in
$\boldsymbol K$. There are several other ways to perform this
computation, such as Gaussian function and concatenation. Dot
product is shown to be the simplest but most effective
one~\cite{wang2018non}. In this work, we use dot product as the
similarity function. In general, we can apply linear transformations
on input matrices, leading to following attention operator:
\begin{equation}
\begin{aligned}
\boldsymbol E &= (\boldsymbol W^K \boldsymbol K)^T \boldsymbol W^Q \boldsymbol Q &\in& \mathbb{R}^{n \times m},\\
\boldsymbol O &= \boldsymbol W^V \boldsymbol V \mbox{softmax}(\boldsymbol E) &\in& \mathbb{R}^{p' \times m},
\end{aligned}
\end{equation}
where $\boldsymbol W^V \in \mathbb{R}^{p' \times p}$ $\boldsymbol
W^K \in \mathbb{R}^{d' \times d}$ and $\boldsymbol W^Q \in
\mathbb{R}^{d' \times d}$. In the following discussions, we will
skip linear transformations for the sake of notational simplicity.

The computational cost of the attention operator as described in
Eq.~\ref{eq:att} is $O(n\times d \times m) + O(p \times n \times m)
= O(n \times m \times (d + p))$. The space complexity for storing
intermediate coefficient matrix $\boldsymbol E$ is $O(n \times m)$.
If $d = p$ and $m = n$, the time and space complexities are
$O(m^2\times d)$ and $O(m^2)$, respectively.

\subsection{Hard Attention Operator}

The attention operator described above uses soft attention, since
responses to each query vector $\boldsymbol q_i$ are calculated by
taking weighted sum over all value vectors. In contrast, hard
attention operator~\cite{xu2015show} only selects a subset of key
and value vectors for computation. Suppose $k$ key vectors~($k < n$)
are selected from the input matrix $\boldsymbol K$ and the indices
are $i_1, i_2, \cdots, i_k$ with $i_m < i_n$ and $1 \le m < n \le
k$. With selected indices, new key and value matrices are
constructed as $\boldsymbol{\tilde K} = [\mathbf{k}_{i_1},
\mathbf{k}_{i_2}, \ldots, \mathbf{k}_{i_k}] \in \mathbb{R}^{d\times
k}$ and $\boldsymbol{\tilde V} = [\mathbf{v}_{i_1},
\mathbf{v}_{i_2}, \ldots, \mathbf{v}_{i_k}] \in \mathbb{R}^{p\times
k}$. The output of the hard attention operator is obtained by
$\boldsymbol O =
\mbox{attn}(\boldsymbol Q, \boldsymbol{\tilde K}, \boldsymbol{\tilde V})$.
The hard attention operator is converted into a stochastic process
in~\cite{xu2015show} by setting $k$ to 1 and use probabilistic
sampling. For each query vector, it only selects one value vector by
probabilistic sampling based on normalized similarity scores given
by $\mbox{softmax}(\boldsymbol K \boldsymbol q_i)$. The hard
attention operators using probabilistic sampling are not
differentiable, and requires reinforcement learning techniques for
training. This makes soft attention more popular for easier
back-propagation training~\cite{ling2017coarse}.

By attending to less key vectors, the hard attention operator is
computationally more efficient than the soft attention operator. The
time and space complexities of the hard attention operator are $O(m
\times k \times d)$ and $O(m\times k)$, respectively. When $k \ll
m$, the hard attention operator reduces time and space complexities
by a factor of $m$ compared to the soft attention operator. Besides
computational efficiency, the hard attention operator is shown to
have better performance than the soft attention
operator~\cite{xu2015show,luong2015effective}, because it only selects
important feature vectors to
attend~\cite{malinowski2018learning,juefei2016deepgender}.

\subsection{Graph Attention Operator}

The graph attention operator~(GAO) was proposed
in~\cite{velivckovic2017graph}, and it applies the soft attention
operator on graph data. For each node in a graph, it attends to its
neighboring nodes. Given a graph with $N$ nodes, each with $d$
features, the layer-wise forward propagation operation of GAO
in~\cite{velivckovic2017graph} is defined as
\begin{equation}
\begin{aligned}
\boldsymbol {\tilde E} &= (\boldsymbol X^T \boldsymbol X) \circ \boldsymbol A,\\
\boldsymbol O &= \boldsymbol X \mbox{softmax}(\boldsymbol{\tilde E}),
\end{aligned}
\end{equation}
where $\circ$ denotes element-wise matrix multiplication, $\boldsymbol A \in
\{0,1\}^{N\times N}$ and $\boldsymbol X = [\boldsymbol x_1, \boldsymbol x_2,
\ldots, \boldsymbol x_N] \in \mathbb{R}^{d\times N}$ are the adjacency and
feature matrices of a graph. Each $\boldsymbol x_i \in \mathbb{R}^{d}$ is node
$i$'s feature vector. In some situations, $\boldsymbol A$ can be normalized as
needed~\cite{kipf2016semi}. Note that the softmax function only applies to
nonzero elements of $\boldsymbol{\tilde E}$.

The time complexity of GAO is $O(Cd)$, where $C$ is
the number of edges. On a dense graph with $C \approx N^2$, this
reduces to $O(N^2d)$. On a sparse graph, sparse matrix operations
are required to compute GAO with this efficiency. However, current
tensor manipulation frameworks such as TensorFlow do not support
efficient batch training with sparse matrix
operations~\cite{velivckovic2017graph}, making it hard to achieve
this efficiency. In general, GAO consumes excessive computational
resources, preventing its applications on large graphs.


\section{Hard and Channel-Wise Attention Operators and Networks}\label{sec:networks}

In this section, we describe our proposed hard graph attention
operator~(hGAO) and channel-wise graph attention operator~(cGAO).
hGAO applies the hard attention operation on graph data, thereby
saving computational cost and improving performance. cGAO performs
attention operation on channels, which avoids the dependency on
the adjacency matrix and significantly improves efficiency
in terms of computational resources. Based on these operators, we
propose the deep graph attention networks for network embedding
learning.

\subsection{Hard Graph Attention Operator}

Graph attention operator~(GAO) consumes excessive computation
resources, including computational cost and memory usage, when
graphs have a large number of nodes, which is very common in
real-world applications. Given a graph with $N$ nodes, each with $d$
features, GAO requires $O(N^2d)$ and $O(N^2)$ time and space
complexities to compute its outputs. This means the computation cost
and memory required grow quadratically in terms of graph size. This
prohibits the application of GAO on graphs with a large number of
nodes. In addition, GAO uses the soft attention mechanism, which
computes responses of each node from all neighboring nodes in the
graph. Using hard attention operator to replace the soft attention
operator can reduce computational cost and improve learning
performance. However, there is still no hard attention operator on
graph data to the best of our knowledge. Direct use of the hard
attention operator as in~\cite{xu2015show} on graph data still
incurs excessive computational resources. It requires the
computation of the normalized similarity scores for probabilistic
sampling, which is the key factor of high requirements on
computational resources.

To address the above limitations of GAO, we propose the hard graph
attention operator~(hGAO) that applies hard attention on graph data
to save computational resources. For all nodes in a graph, we use a
projection vector $\boldsymbol p \in \mathbb{R}^{d}$ to select the
$k$-most important nodes to attend. Following the notations defined
in Section~\ref{sec:relatedwork}, the layer-wise forward propagation
function of hGAO is defined as
\begin{align}
&\boldsymbol y = \frac{|\boldsymbol X^T \boldsymbol p|}{\lVert \boldsymbol p \rVert} &\in& \mathbb{R}^{N}\label{eq:proj} \\
&\mbox{for } i = 1, 2, \cdots, N \mbox{ do} \nonumber \\
&\,\,\,\,\,\,\,\,\boldsymbol{idx}_i = \mbox{Ranking}_k(\boldsymbol{A_{:i}} \circ \boldsymbol y) &\in& \mathbb{R}^{k}\label{eq:rank} \\
&\,\,\,\,\,\,\,\,\boldsymbol{\hat X}_i = \boldsymbol X(:,\boldsymbol{idx}_i) &\in& \mathbb{R}^{d\times k}\label{eq:extract} \\
&\,\,\,\,\,\,\,\,\boldsymbol{\tilde y}_i = \mbox{sigmoid} (\boldsymbol y(\boldsymbol{idx}_i)) &\in& \mathbb{R}^{k}\label{eq:sigmoid}\\
&\,\,\,\,\,\,\,\,\boldsymbol{\tilde X}_i = \boldsymbol{\hat X}_i \mbox{diag}(\boldsymbol{\tilde y}_i) &\in& \mathbb{R}^{d\times k}\label{eq:gate} \\
&\,\,\,\,\,\,\,\,\boldsymbol z_i = \mbox{attn}(\boldsymbol x_i, \boldsymbol{\tilde X}_i, \boldsymbol{\tilde X}_i) &\in& \mathbb{R}^{d}\label{eq:iatt} \\
& \boldsymbol Z = [\boldsymbol z_1, \boldsymbol z_2, \ldots, \boldsymbol z_N] & \in& \mathbb{R}^{d \times N}\label{eq:combine}
\end{align}
where $\boldsymbol{A_{:i}}$ denotes the $i$th column of matrix
$\boldsymbol{A}$, $\boldsymbol X(:,\boldsymbol{idx}_i)$ contains a
subset of columns of $\boldsymbol X$ indexed by
$\boldsymbol{idx}_i$, $|\cdot|$ computes element-wise absolute
values, $\circ$ denotes element-wise matrix/vector multiplication,
$\mbox{diag}(\cdot)$ constructs a diagonal matrix with the input
vector as diagonal elements, $\mbox{Ranking}_k(\cdot)$ is an
operator that performs the $k$-most important nodes selection for
the query node $i$ to attend and is described in detail below.

We propose a novel node selection method in hard attention. For each
node in the graph, we adaptively select the $k$ most important
adjacent nodes. By using a trainable projection vector $\boldsymbol
p$, we compute the absolute scalar projection of $\boldsymbol X$ on
$\boldsymbol p$ in Eq.~(\ref{eq:proj}), resulting in $\boldsymbol y
= [y_1, y_2, \cdots, y_N]^T$. Here, each $y_i$ measures the
importance of node $i$. For each node $i$, the
$\mbox{Ranking}_k(\cdot)$ operation in Eq.~(\ref{eq:rank}) ranks
node $i$'s adjacent nodes by their projection values in $\boldsymbol
y$, and selects nodes with the $k$ largest projection values.
Suppose the indices of selected nodes for node $i$ are
$\boldsymbol{idx}_i = [i_1, i_2, \cdots, i_k]$, node $i$ attends to
these $k$ nodes, instead of all adjacent nodes. In
Eq.~(\ref{eq:extract}), we extract new feature vectors
$\boldsymbol{\hat X}_i = [\boldsymbol x_{i_1}, \boldsymbol{x}_{i_2},
\ldots, \boldsymbol{x}_{i_k}] \in \mathbb{R}^{d\times k}$ using the
selected indices $\boldsymbol{idx}_i$. Here, we propose to use a
gate operation to control information flow. In
Eq.~(\ref{eq:sigmoid}), we obtain the gate vector
$\boldsymbol{\tilde y}$ by applying the sigmoid function to the
selected scalar projection values $\boldsymbol
y(\boldsymbol{idx}_i)$. By matrix multiplication $\boldsymbol{\hat
X}_i \mbox{diag}(\boldsymbol{\tilde y}_i)$ in Eq.~(\ref{eq:gate}),
we control the information of selected nodes and make the projection
vector $\boldsymbol p$ trainable with gradient back-propagation. We
use attention operator to compute the response of node $i$ in
Eq.~(\ref{eq:iatt}). Finally, we construct the output feature matrix
$\boldsymbol Z$ in Eq.~(\ref{eq:combine}). Note that the projection
vector $\boldsymbol p$ is shared across all nodes in the graph. This
means hGAO only involves $d$ additional parameters, which may not
increase the risk of over-fitting.

\begin{figure*}[t] \includegraphics[width=\textwidth]{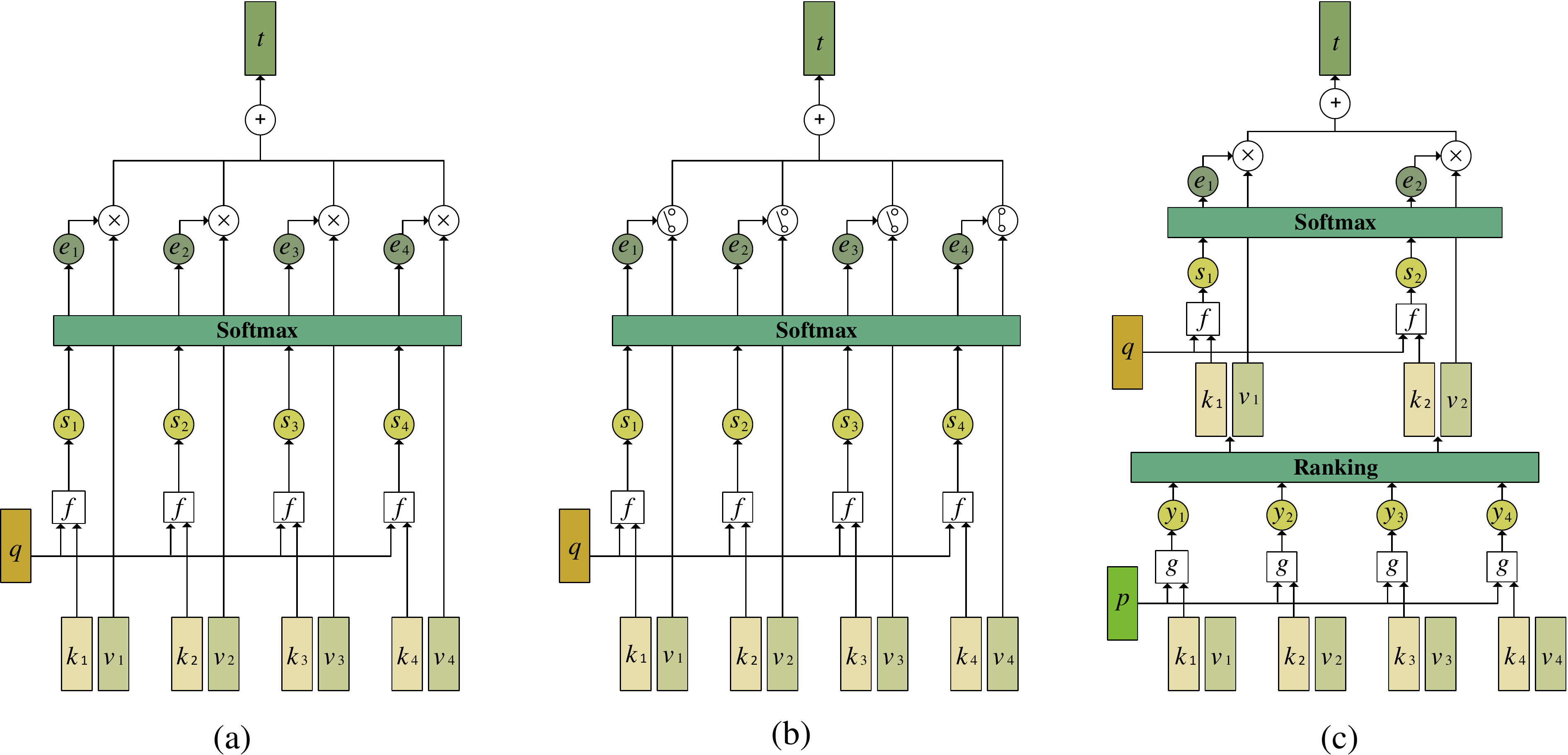}
\caption{Illustration of GAO~(a), hard attention operator
in~\cite{xu2015show}~(b), and our proposed hGAO~(c). $\boldsymbol q$
is the feature vector of a node with four neighboring nodes in a
graph. $\boldsymbol k_i$ and $\boldsymbol v_i$ are key and value
vectors of the neighboring node $i$. In GAO~(a), similarity scores
are computed between query vector and key vectors, leading to scalar
values $s_i$. The softmax normalizes these values and converts them
into weights. The output is computed by taking a weighted sum of
value vectors. In hard attention operator~(b), the output is
generated by probabilistic sampling, which samples a vector from
value vectors using computed weights $e_i$. In hGAO~(c), a projection
vector $\boldsymbol p$ is used to compute the importance scores
$y_i$. Based on these importance scores, two out of four nodes are
selected by ranking. The output is computed by applying soft
attention on selected nodes.} \label{fig:atts}
\end{figure*}

By attending to less nodes in graphs, hGAO is computationally more
efficient than GAO. The time complexity of hGAO is $O(N \times \log
N \times k + N \times k \times d^2)$ if using max heap for
$k$-largest selection. When $k \ll N$ and $d \ll N$, hGAO consumes
less time compared to the GAO. The space complexity of hGAO is
$O(N^2)$ since we need to store the intermediate score matrix during
$k$-most important nodes selection. Besides computational
efficiency, hGAO is expected to have better performance than GAO,
because it selects important neighboring nodes to
attend~\cite{malinowski2018learning}. We show in our experiments
that hGAO outperforms GAO, which is consistent with the performance
of hard attention operators in NLP and computation vision
fields~\cite{xu2015show,luong2015effective}.

This method can be considered as a trade-off between soft attention
and the hard attention in~\cite{xu2015show}. The query node attends
all neighboring nodes in soft attention operators. In the hard
attention operator in~\cite{xu2015show}, the query node attends to
only one node that is probabilistically sampled from neighboring
nodes based on the coefficient scores. In our hGAO, we employ an
efficient ranking method to select $k$-most important neighboring
nodes for the query node to attend. This avoids computing the
coefficient matrix and reduces computational cost. The proposed gate
operation enables training of the projection vector $\boldsymbol p$
using back-propagation~\cite{lecun2012efficient}, thereby avoiding
the need of using reinforcement learning
methods~\cite{rao2017attention} for training as
in~\cite{xu2015show}. Figure~\ref{fig:atts} provides illustrations
and comparisons among soft attention operator, hard attention
in~\cite{xu2015show}, and our proposed hGAO.

Another possible way to compute the hard attention operator as hGAO
is to implement the $k$-most important node selection based on the
coefficient matrix. For each query node, we can select $k$
neighboring nodes with $k$-largest similarity scores. The responses
of the query node is calculated by attending to these $k$ nodes.
This method is different from our hGAO in that it needs to compute
the coefficient matrix, which takes $O(N^2 \times d)$ time
complexity. The hard attention operator using this implementation
consumes much more computational resources than hGAO. In addition,
the selection process in hGAO employs a trainable projection vector
$\boldsymbol p$ to achieve important node selection. Making the
projection vector $\boldsymbol p$ trainable allows for the learning
of importance scores from data.

\subsection{Channel-Wise Graph Attention Operator}

The proposed hGAO computes the hard attention operator on graphs
with reduced time complexity, but it still incurs the same space
complexity as GAO. At the same time, both GAO and hGAO need to use
the adjacency matrix to identify the neighboring nodes for the query
node in the graph. Unlike grid like data such as images and texts,
the number and ordering of neighboring nodes in a graph are not
fixed. When performing attention operations on graphs, we need to
rely on the adjacency matrix, which causes additional usage of
computational resources. To further reduce the computational
resources required by attention operators on graphs, we propose the
channel-wise graph attention operator, which gains significant
advantages over GAO and hGAO in terms of computational resource
requirements.

\begin{figure*}[t]
\includegraphics[width=\textwidth]{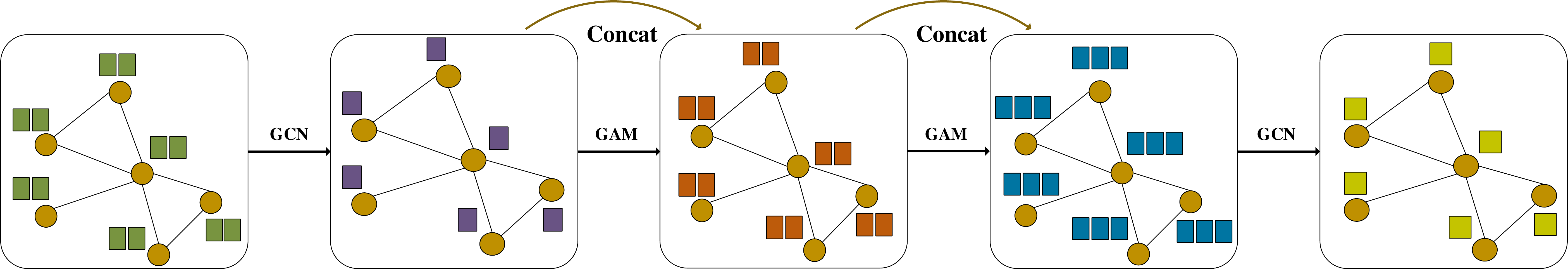}
\caption{An illustration of our proposed GANet described in Section~\ref{sec:GHAN}.
In this example, the input graph contains 6 nodes, each of which has two features.
A GCN layer is used to transform input feature vectors into low-dimensional representations.
After that, we stack two GAMs for feature extraction. To facilitate feature reuse
and gradients back-propagation, we add skip concatenation connections for GAMs. Finally,
a GCN layer is used to output designated number of feature maps, which can be
directly used for node classification predictions or used as inputs of following operations.
} \label{fig:architecture}
\end{figure*}

Both GAO and our hGAO use the node-wise attention mechanism in which
the output feature vector of node $i$ is obtained by attending the
input feature vector to all or selected neighboring nodes. Here, we
propose to perform attention operation from the perspective of
channels, resulting in our channel-wise graph attention
operator~(cGAO). For each channel $\boldsymbol X_{i:}$, we compute
its responses by attending it to all channels. The layer-wise
forward propagation function of cGAO can be expressed as
\begin{equation}
\begin{aligned}
\boldsymbol E &= \boldsymbol X \boldsymbol X^T  &\in& \mathbb{R}^{d\times d},\\
\boldsymbol O &= \mbox{softmax}(\boldsymbol E) \boldsymbol X &\in&
\mathbb{R}^{d\times N}.
\end{aligned}
\end{equation}
Note that we avoid the use of adjacency matrix $\boldsymbol A$,
which is different from GAO and hGAO. When computing the coefficient
matrix $\boldsymbol E$, the similarity score between two feature
maps $\boldsymbol X_{i:}$ and $\boldsymbol X_{j:}$ are calculated by
$e_{ij} = \sum_{k=1}^{N} X_{ik} \times X_{jk}$. It can be seen that
features within the same node communicate with each other, and there
is no communication among features located in different nodes. This
means we do not need the connectivity information provided by
adjacency matrix $\boldsymbol A$, thereby avoiding the dependency on
the adjacency matrix used in node-wise attention operators. This
saves computational resources related to operations with the
adjacency matrix.

The computational cost of cGAO is $O(Nd^2)$, which is lower than
that of GAO if $d < N$. When applying attention operators on graph
data, we can control the number of feature maps $d$, but it is hard
to reduce the number of nodes in graphs. On large graphs with $d \ll
N$, cGAO has computational advantages over GAO and hGAO, since its
time complexity is only linear to the size of graphs. The space
complexity of cGAO is $O(d^2)$, which is independent of graph size.
This means the application of cGAO on large graphs does not suffer
from memory issues, which is especially useful on memory limited
devices such as GPUs and mobile devices. Table~\ref{table:gaos}
provides theoretical comparisons among GAO, hGAO and cGAO in terms
of the time and space complexities. Therefore, cGAO enables
efficient parallel training by removing the dependency on the
adjacency matrix in graphs and significantly reduces the usage of
computational resources.

\begin{table}[t]
\caption{Comparison of time and space complexities among GAO, hGAO,
and cGAO.} \label{table:gaos}
\begin{tabularx}{\columnwidth}{  lx{3.5cm} cx{4cm} Y }
    \hline
    \textbf{Operator} & \textbf{Time~Complexity} & \textbf{Space~Complexity} \\ \hline\hline
    GAO    & $O(N^2\times d)$ & $O(N^2)$   \\ \hline
    hGAO   & $O(N \times \log N \times k + N \times k \times d^2)$  & $O(N^2)$    \\ \hline
    cGAO   & $O(N \times d^2)$  & $O(d^2)$    \\ \hline
    \hline
\end{tabularx}
\end{table}

\subsection{The Proposed Graph Attention Networks}\label{sec:GHAN}

To use our hGAO and cGAO, we design a basic module known as the
graph attention module~(GAM). The GAM consists of two operators;
those are, a graph attention operator and a graph convolutional
network (GCN) layer~\cite{kipf2016semi}. We combine these two
operators to enable efficient information propagation within graphs.
For GAO and hGAO, they aggregate information from neighboring nodes
by taking weighted sum of feature vectors from adjacent nodes. But
there exists a situation that weights of some neighboring nodes are
close to zero, preventing the information propagation of these
nodes. In cGAO, the attention operator is applied among channels and
does not involve information propagation among nodes. To overcome
this limitation, we use a GCN layer, which applies the same weights
to neighboring nodes and aggregate information from all adjacent
nodes. Note that we can use any graph attention operator such as
GAO, hGAO and cGAO. To facilitate feature reuse and gradients
back-propagation, we add a skip connection by concatenating inputs
and outputs of the GCN layer.

Based on GAM, we design graph attention networks, denoted as GANet,
for network embedding learning. In GANet, we first apply a GCN
layer, which acts as a graph embedding layer to produce
low-dimensional representations for nodes. In some data like
citation networks dataset~\cite{kipf2016semi}, nodes usually have
very high-dimensional feature vectors. After the GCN layer, we stack
multiple GAMs depending on the complexity of the graph data. As each
GAM only aggregates information from neighboring nodes, stacking
more GAMs can collect information from a larger parts of the graph.
Finally, a GCN layer is used to produce designated number of output
feature maps. The outputs can be directly used as predictions of
node classification tasks. We can also add more operations to
produce predictions for graph classification tasks.
Figure~\ref{fig:architecture} provides an example of our GANet.
Based on this network architecture, we denote the networks using
GAO, hGAO and cGAO as GANet, hGANet and cGANet, respectively.

\begin{table*}[t]
\centering \caption{Statistics of datasets used in graph classification tasks under inductive
learning settings. We use the D\&D, PROTEINS, COLLAB, MUTAG, PTC, and IMDB-M datasets.
}\label{table:inducdatasets}
\begin{tabularx}{\textwidth}{  XYYYYYYY }
    \hline
    \textbf{Dataset} & \textbf{Total~Graphs} & \textbf{Train~Graphs} & \textbf{Test~Graphs} &
    \textbf{Nodes~(max)} & \textbf{Nodes~(avg)} & \textbf{Degree} & \textbf{Classes} \\ \hline\hline
    MUTAG     & 188  & 170  & 18   & 28   & 17.93   & 2.19  & 2  \\ \hline
    PTC       & 344  & 310  & 34   & 109  & 25.56   & 1.99  & 2  \\ \hline
    PROTEINS  & 1113 & 1002 & 111  & 620  & 39.06   & 3.73  & 2  \\ \hline
    D\&D      & 1178 & 1061 & 117  & 5748 & 284.32  & 4.98  & 2  \\ \hline
    IMDB-M    & 1500 & 1350 & 150  & 89   & 13.00   & 10.14 & 3  \\ \hline
    COLLAB    & 5000 & 4500 & 500  & 492  & 74.49   & 65.98 & 3  \\ \hline
    \hline
\end{tabularx}
\end{table*}

\begin{table*}[t]
\centering \caption{Statistics of datasets used in node classification tasks under transductive
learning settings. We use the Cora, Citeseer, and Pubmed datasets.}\label{table:transdatasets}
\begin{tabularx}{\textwidth}{  XYYYYYYYY }
    \hline
    \textbf{Dataset} & \textbf{Nodes} &
    \textbf{Features} & \textbf{Training} &
    \textbf{Validation} & \textbf{Testing} & \textbf{Degree} & \textbf{Classes} \\ \hline\hline
    Cora      & 2708   & 1433 & 140   & 500 & 1000 & 4 & 7    \\ \hline
    Citeseer  & 3327   & 3703 & 120   & 500 & 1000 & 5 & 6    \\ \hline
    Pubmed    & 19717  & 500  & 60    & 500 & 1000 & 6 & 3    \\ \hline
    \hline
\end{tabularx}
\end{table*}


\section{Experimental Studies}

In this section, we evaluate our proposed graph attention networks
on node classification and graph classification tasks. We first
compare our hGAO and cGAO with GAO in terms of computation resources
such as computational cost and memory usage. Next, we compare our
hGANet and cGANet with prior state-of-the-art models under inductive
and transductive learning settings. Performance studies among GAO,
hGAO, and cGAO are conducted to show that our hGAO and cGAO achieve
better performance than GAO. We also conduct some performance
studies to investigate the selection of some hyper-parameters.

\subsection{Datasets}\label{sec:dataset}

We conduct experiments on graph classification tasks under inductive
learning settings and node classification tasks under transductive
learning settings. Under inductive learning settings, training and
testing data are separate. The test data are not accessible during
training time. The training process will not learn about graph
structures of the test data. For graph classification tasks under
inductive learning settings, we use the
MUTAG~\cite{niepert2016learning}, PTC~\cite{niepert2016learning},
PROTEINS~\cite{borgwardt2005protein},
D\&D~\cite{dobson2003distinguishing},
IMDB-M~\cite{yanardag2015structural}, and
COLLAB~\cite{yanardag2015structural} datasets to fully evaluate our
proposed methods. MUTAG, PTC, PROTEINS and D\&D are four
benchmarking bioinformatics datasets. MUTAG and PTC are much smaller
than PROTEINS and D\&D in terms of number of graphs and average
nodes in graphs. Compared to large datasets, evaluations on small
datasets can help investigate the risk of over-fitting, especially
for deep learning based methods. COLLAB, IMDB-M are two social
network datasets. For these datasets, we follow the same settings as
in~\cite{zhang2018end}, which employs 10-fold cross
validation~\cite{chang2011libsvm} with 9 folds for training and 1
fold for testing. The statistics of these datasets are summarized in
Table~\ref{table:inducdatasets}.

Unlike inductive learning settings, the unlabeled data and graph structure are
accessible during the training process under transductive learning settings.
To be specific, only a small part of nodes in the graph are labeled while the
others are not. For node classification tasks under transductive learning
settings, we use three benchmarking datasets; those are
Cora~\cite{sen2008collective}, Citeseer, and Pubmed~\cite{kipf2016semi}. These
datasets are citation networks. Each node in the graph represents a document
while an edge indicates a citation relationship. The graphs in these datasets
are attributed and the feature vector of each node is generated by bag-of-word
representations. The dimensions of feature vectors of three datasets are
different depending on the sizes of dictionaries. Following the same
experimental settings in~\cite{kipf2016semi}, we use 20 nodes, 500 nodes, and
500 nodes for training, validation, and testing, respectively.

\begin{table*}[t]
\caption{
Comparison of results among GAO, hGAO, and cGAO on different graph sizes in terms of
the number of MAdd, memory usage, and CPU prediction time. The input sizes are describe
by ``number of nodes $\times$ number of features''. The prediction time is the total
execution time on CPU.}
\label{table:layers_cmp}
\begin{tabularx}{\textwidth}{c lx{2.4cm}  cx{2.4cm} cx{2.2cm} Y Y Y Y}
    \hline
    \textbf{Input} & \textbf{Layer} & \textbf{MAdd} & \textbf{Cost Saving} & \textbf{Memory} & \textbf{Memory Saving} & \textbf{Time} & \textbf{Speedup} \\ \hline\hline
    \multirow{3}{*}{$1000\times48$}
     & GAO     & 100.61m   & 0.00\%   & 4.98MB   &  0.00\%     & 8.19ms       & 1.0$\times$ \\ 
     & hGAO    & 37.89m  & 62.34\%   & 4.98MB   &  0.00\%     & 5.61ms       & 1.46$\times$ \\ 
     & cGAO    & \textbf{9.21m} & \textbf{90.84\%}  & \textbf{0.99MB} & \textbf{80.12\%}  & \textbf{0.82ms} & \textbf{9.99$\times$} \\ \hline
     \multirow{3}{*}{$10000\times48$}
     & GAO     & 9,646.08m   & 0.00\%   & 409.6MB  &  0.00\%     & 947.24ms      & 1.0$\times$  \\ 
     & hGAO    & 468.96m   & 95.14\%  & 409.6MB  &  0.00\%     & 371.12ms       & 2.55$\times$ \\ 
     & cGAO    & \textbf{92.16m} & \textbf{99.04\%} & \textbf{9.61MB} & \textbf{97.65\%} & \textbf{17.96ms} & \textbf{52.74$\times$} \\ \hline
    \multirow{3}{*}{$20000\times48$}
     & GAO     & 38,492.16m  & 0.00\%   & 1,619.2MB  & 0.00\%     & 12,784.45ms   & 1.0$\times$ \\ 
     & hGAO    & 1,137.97m   & 97.04\%  & 1,619.2MB  & 0.00\%     & 4,548.62ms      & 2.81$\times$ \\ 
     & cGAO    & \textbf{184.32m} & \textbf{99.52\%} & \textbf{19.2MB} & \textbf{98.81\%} & \textbf{29.71ms} & \textbf{430.31$\times$} \\ \hline
    \hline
\end{tabularx}
\end{table*}

\begin{table*}[t]
\centering \caption{Comparison of results of graph classification experiments with prior state-of-the-art
models in terms of accuracies on the
D\&D, PROTEINS, COLLAB, MUTAG, PTC, and IMDB-M datasets. ``-'' denotes the result not available.}
\label{table:induc}
\begin{tabularx}{\textwidth}{  X  YYYYYY }
    \hline
    \textbf{Models} & \textbf{D\&D} & \textbf{PROTEINS} & \textbf{COLLAB} & \textbf{MUTAG} & \textbf{PTC} & \textbf{IMDB-M} \\ \hline\hline
    GRAPHSAGE~\cite{hamilton2017inductive}      & 75.42\% & 70.48\%  & 68.25\%  & - & - & -  \\ \hline
    PSCN~\cite{niepert2016learning}             & 76.27\% & 75.00\%  & 72.60\%  & 88.95\% & 62.29\% & 45.23\%  \\ \hline
    SET2SET~\cite{vinyals2015order}             & 78.12\% & 74.29\%  & 71.75\%  & - & - & -  \\ \hline
    DGCNN~\cite{zhang2018end}                   & 79.37\% & 76.26\%  & 73.76\%  & 85.83\% & 58.59\% & 47.83\%  \\ \hline
    DiffPool~\cite{ying2018hierarchical}        & 80.64\% & 76.25\%  & 75.48\%  & - & - & - \\ \hline
    cGANet      & 80.86\% & 78.23\%  & 76.96\%  & 89.00\%  & 63.53\% & 48.93\% \\ \hline
    \textbf{hGANet}                      & \textbf{81.71\%}   & \bf 78.65\% & \bf 77.48\%
                                                & \textbf{90.00\%}   & \textbf{65.02\%} & \textbf{49.06\%} \\ \hline
    \hline
\end{tabularx}
\end{table*}

\subsection{Experimental Setup}\label{sec:expsetup}

In this section, we describe the experimental setup for inductive
learning and transductive learning tasks. For inductive learning
tasks, we adopt the model architecture of DGCNN~\cite{zhang2018end}.
DGCNN consists of four parts; those are graph convolution layers,
soft pooling, 1-D convolution layers and dense layers. We replace
graph convolution layers with our hGANet described in
Section~\ref{sec:GHAN} and the other parts remain the same. The
hGANet contains a starting GCN layer, four GAMs and an ending GCN
layer. Each GAM is composed of a hGAO, and a GCN layer. The starting
GCN layer outputs 48 feature maps. Each hGAO and GCN layer within
GAMs outputs 12 feature maps. The final GCN layer produces 97
feature maps as the original graph convolution layers in DGCNN. The
skip connections using concatenation is employed between the input
and output feature maps of each GAM. The hyper-parameter $k$ is set
to 8 in each hGAO, which means each node in a graph selects 8 most
important neighboring nodes to compute the response. We apply
dropout~\cite{srivastava2014dropout} with the keep rate of 0.5 to
the feature matrix in every GCN layer. For experiments on cGANet,
we use the same settings.

For transductive learning tasks, we use our hGANet to perform node
classification predictions. Since the feature vectors for nodes are generated using the
bag-of-words method, they are high-dimensional sparse features. The first GCN
layer acts like an embedding layer to reduce them into low-dimensional
features. To be specific, the first GCN layer outputs 48 feature maps to
produce 48 embedding features for each node. For different datasets, we stack
different number of GAMs. Specifically, we use 4, 2, and 3 GAMs for Cora,
Citeseer, and Pubmed, respectively. Each hGAO and GCN layer in GAMs outputs
16 feature maps. The last GCN layer produces the prediction on each node in
the graph. We apply dropout with the keep rate of 0.12 on feature matrices in
each layer. We also set $k$ to 8 in all hGAOs. We employ identity
activation function as~\cite{gao2018large} for all layers in the model. To
avoid over-fitting, we apply $L_2$ regularization with $\lambda=0.0001$. All
trainable weights are initialized with Glorot
initialization~\cite{glorot2010understanding}. We use Adam
optimizer~\cite{kingma2014adam} for training.

\subsection{Comparison of Computational Efficiency}

According to the theoretical analysis in Section~\ref{sec:networks}, our
proposed hGAO and cGAO have efficiency advantages over GAO in terms of the
computational cost and memory usage. The advantages are expected to be more
obvious as the increase of the number of nodes in a graph. In this section, we
conduct simulated experiments to evaluate these theoretical analysis results.
To reduce the influence of external factors, we use the network with a single
graph attention operator and apply TensorFlow profile
tool~\cite{abadi2016tensorflow} to report the number of multiply-adds~(MAdd),
memory usage, and CPU inference time on simulated graph data.

The simulated data are create with the shape of ``number of nodes $\times$
number of feature maps''. For all simulated experiments, each node on the
input graph has 48 features. We test three graph sizes; those are 1000,
1,0000, and 20,000, respectively. All tested graph operators output 48 feature
maps including GAO, hGAO, and cGAO. For hGAOs, we set $k=8$ in all
experiments, which is the value of hyper-parameter $k$ tuned on graph
classification tasks. We report the number of MAdd, memory usage, and CPU
inference time.

The comparison results are summarized in Table~\ref{table:layers_cmp}. On the
graph with 20,000 nodes, our cGAO and hGAO provide 430.31 and 2.81 times
speedup compared to GAO. In terms of the memory usage, cGAO can save 98.81\%
compared to GAO and hGAO. When comparing across different graph sizes, the
effects of speedup and memory saving are more apparent as the graph size
increases. This is consistent with our theoretical analysis on hGAO and cGAO.
Our hGAO can save computational cost compared to GAO. cGAO achieves great
computational resources reduction, which makes it applicable on large graphs.
Note that the speed up of hGAO over GAO is not as apparent as the
computational cost saving due to the practical implementation limitations.

\begin{table}[t]
\centering \caption{Comparison of results of node classification experiments with
prior state-of-the-art models on the Cora, Citeseer, and Pubmed datasets.}
\label{table:trans}
\begin{tabularx}{\columnwidth}{  X YYY  }
    \hline
    \textbf{Models} & \textbf{Cora} & \textbf{Citeseer} & \textbf{Pubmed} \\ \hline\hline
    DeepWalk~\cite{perozzi2014deepwalk}            & 67.2\% & 43.2\%  & 65.3\%   \\ \hline
    Planetoid~\cite{yang2016revisiting}            & 75.7\% & 64.7\%  & 77.2\%   \\ \hline
    Chebyshev~\cite{defferrard2016convolutional}   & 81.2\% & 69.8\%  & 74.4\%   \\ \hline
    GCN~\cite{kipf2016semi}                        & 81.5\% & 70.3\%  & 79.0\%   \\ \hline
    GAT~\cite{velivckovic2017graph}                & 83.0 $\pm$ 0.7\% & 72.5 $\pm$ 0.7\% & 79.0 $\pm$ 0.3\% \\ \hline
    \textbf{hGANet}                             & \textbf{83.5 $\pm$ 0.7\%}
                                                & \textbf{72.7 $\pm$ 0.6\%}
                                                & \textbf{79.2 $\pm$ 0.4\%} \\ \hline
    \hline
\end{tabularx}
\end{table}

\subsection{Results on Inductive Learning Tasks}

\begin{table*}[t]
\caption{
Comparison of results of graph classification experiments among GAO, hGAO, and cGAO
in terms of accuracies on the
D\&D, PROTEINS, COLLAB, MUTAG, PTC, and IMDB-M datasets. ``OOM'' denotes out of memory.}
\label{table:gats}
\begin{tabularx}{\textwidth}{  XYYYYYY  }
    \hline
    \textbf{Models} & \textbf{D\&D} & \textbf{PROTEINS} & \textbf{COLLAB} & \textbf{MUTAG} & \textbf{PTC} & \textbf{IMDB-M} \\ \hline\hline
    GANet       & OOM & 77.92\%  & 76.06\%  & 87.22\%  & 62.94\% & 48.89\%  \\ \hline
    cGANet      & 80.86\% & 78.23\%  & 76.96\%  & 89.00\%  & 63.53\% & 48.93\% \\ \hline
    hGANet      & \textbf{81.71\%} & \textbf{78.65\%}  & \textbf{77.48\%}  & \textbf{90.00\%}
                & \textbf{65.02\%} & \textbf{49.06\%} \\ \hline
    \hline
\end{tabularx}
\end{table*}

We evaluate our methods on graph classification tasks under inductive learning
settings. To compare our proposed cGAOs with hGAO and GAO, we replace hGAOs
with cGAOs in hGANet, denoted as cGANet. We compare our models with prior
sate-of-the-art models on D\&D, PROTEINS, COLLAB, MUTAG, PTC, and IMDB-M
datasets, which serve as the benchmarking datasets for graph classification
tasks. The results are summarized in Table~\ref{table:induc}.

From the results, we can observe that the our hGANet consistently outperforms
DiffPool~\cite{ying2018hierarchical} by margins of 0.90\%, 1.40\%, and 2.00\%
on D\&D, PROTEINS, and COLLAB datasets, which contain relatively big graphs in
terms of the average number of nodes in graphs. Compared to DGCNN, the
performance advantages of our hGANet are even larger. The superior
performances on large benchmarking datasets demonstrate that our proposed
hGANet is promising since we only replace graph convolution layers in DGCNN.
The performance boosts over the DGCNN are consistently and significant, which
indicates the great capability on feature extraction of hGAO compared to GCN
layers.

On datasets with smaller graphs, our GANets outperform prior state-of-the-art
models by margins of 1.05\%, 2.71\%, and 1.23\% on MUTAG, PTC, and IMDB-M
datasets. The promising performances on small datasets prove that our methods
improve the ability of high-level feature extraction without incurring the
problem of over-fitting. cGANet outperforms prior state-of-the-art models but
has lower performances than hGANet. This indicates that cGAO is also effective
on feature extraction but not as powerful as hGAO. The attention on only
important adjacent nodes incurred by using hGAOs helps to improve the
performance on graph classification tasks.

\subsection{Results on Transductive Learning Tasks}

Under transductive learning settings, we evaluate our methods on node
classification tasks. We compare our hGANet with prior state-of-the-art models
on Cora, Citeseer, and Pubmed datasets in terms of the node classification
accuracy. The results are summarized in Table~\ref{table:trans}. From the
results, we can observe that our hGANet achieves consistently better
performances than GAT, which is the prior state-of-the-art model using graph
attention operator. Our hGANet outperforms GAT~\cite{velivckovic2017graph} on
three datasets by margins of 0.5\%, 0.2\%, and 0.2\%, respectively. This
demonstrates that our hGAO has performance advantage over GAO by attending
less but most important adjacent nodes, leading to better generalization and
performance.

\subsection{Comparison of cGAO and hGAO with GAO}

Besides comparisons with prior state-of-the-art models, we conduct experiments
under inductive learning settings to compare our hGAO and cGAO with GAO. To be
fair, we replace all hGAOs with GAOs in hGANet employed on graph
classification tasks, which results in GANet. GAOs output the same number of
feature maps as the corresponding hGAOs. Like hGAOs, we apply linear
transformations on key and value matrices. This means GANets have nearly the
same number of parameters with hGANets, which additionally contain limited
number of projection vectors in hGAOs. We adopt the same experimental setups
as hGANet. We compare our hGANet and cGANet with GANet on all six datasets for
graph classification tasks described in Section~\ref{sec:dataset}. The
comparison results are summarized in Table~\ref{table:gats}.

The results show that our cGAO and hGAO have significantly better performances
than GAO. Notably, GANet runs out of memory when training on D\&D dataset with
the same experimental setup as hGANet. This demonstrates that hGAO has memory
advantage over GAO in practice although they share the same space complexity.
cGAO outperforms GAO on all six datasets but has slightly lower performances
than hGAO. Considering cGAO dramatically saves computational resources, cGAO is
a good choice when facing large graphs. Since there is no work that realizes
the hard attention operator in~\cite{xu2015show} on graph data, we do not
provide comparisons with it in this work.

\subsection{Performance Study of $k$ in hGAO}

\begin{figure}[t] \includegraphics[width=\columnwidth]{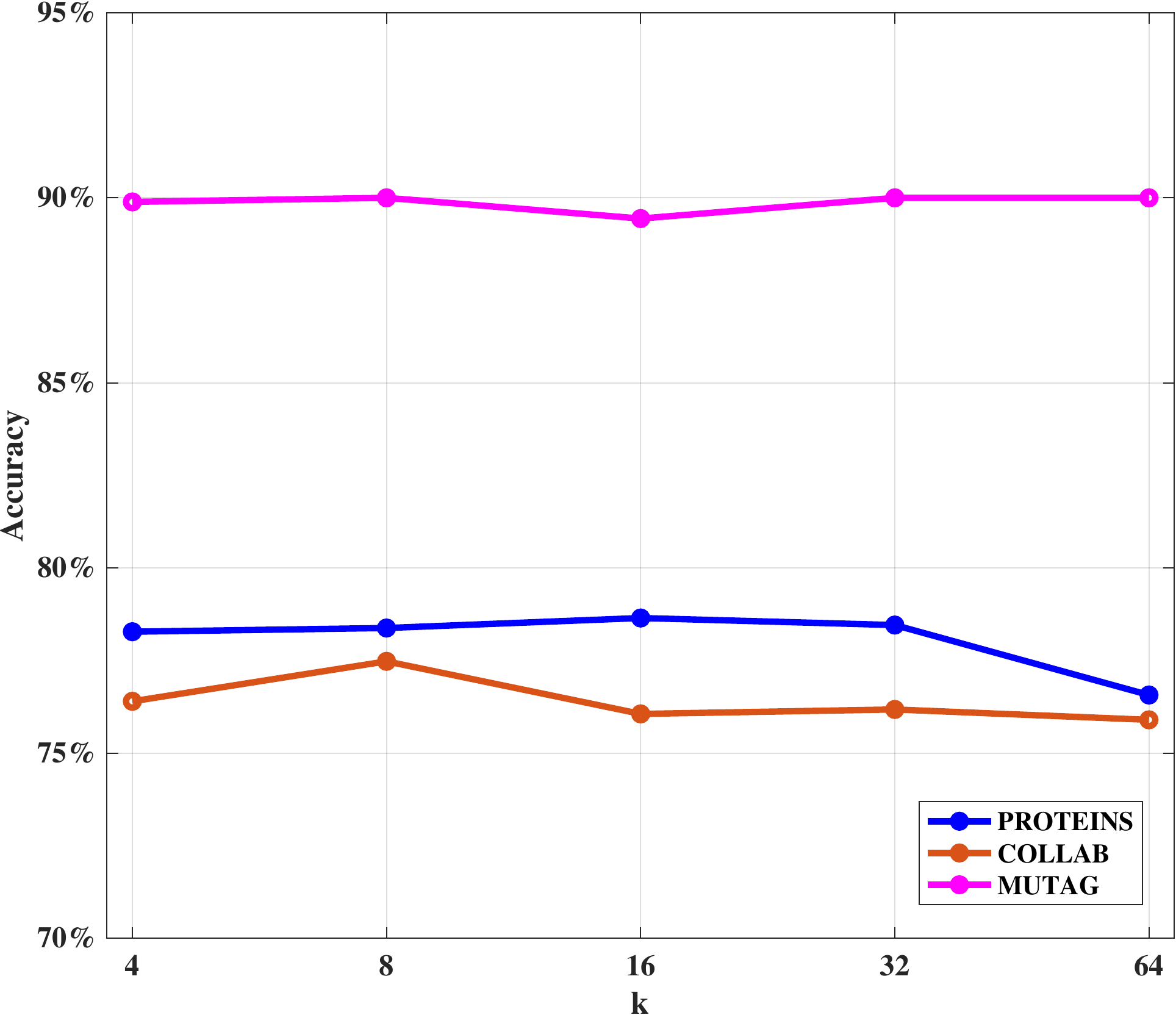}
\caption{ Results of employing different $k$ values in hGAOs using hGANet on
PROTEINS, COLLAB, and MUTAG datasets under inductive learning settings.
We use the same experimental setups
described in Section~\ref{sec:expsetup}. We report the graph classification
accuracies in this figure. We can see that the best performances is achieved
when $k=8$.}\label{fig:k_exp}
\end{figure}

Since $k$ is an important hyper-parameter in hGAO, we conduct experiments to
investigate the impact of different $k$ values on hGANet. Based on hGANet, we
vary the values of $k$ in hGAOs with choices of 4, 8, 16, 32, and 64, which
are reasonable selections for $k$. We report performances of hGANets with
different $k$ values on graph classification tasks on PROTEINS, COLLAB, and
MUTAG datasets, which cover both large and small datasets.

The performance changes of hGANets with different $k$ values are plotted in
Figure~\ref{fig:k_exp}. From the figure, we can see that hGANets achieve the
best performances on all three datasets when $k=8$. The performances start to
decrease as the increase of $k$ values. On PROTEINS and COLLAB datasets, the
performances of hGANets with $k=64$ are significantly lower than those with
$k=8$. This indicates that larger $k$ values make the query node to attend
more adjacent nodes in hGAO, which leads to worse generalization and
performance.


\section{Conclusions}

In this work, we propose novel hGAO and cGAO which are attention
operators on graph data. hGAO achieves the hard attention operation
by selecting important nodes for the query node to attend. By
employing a trainable projection vector, hGAO selects $k$-most
important nodes for each query node based on their projection
scores. Compared to GAO, hGAO saves computational resources and
attends important adjacent nodes, leading to better generalization
and performance. Furthermore, we propose the cGAO, which performs
attention operators from the perspective of channels. cGAO removes
the dependency on the adjacency matrix and dramatically saves
computational resources compared to GAO and hGAO. Based on our
proposed attention operators, we propose a new architecture that
employs a densely connected design pattern to promote feature reuse.
We evaluate our methods under both transductive and inductive
learning settings. Experimental results demonstrate that our hGANets
achieve improved performance compared to prior state-of-the-art
networks. The comparison between our methods and GAO indicates that
our hGAO achieves significant better performance than GAO. Our cGAO
greatly saves computational resources and makes attention operators
applicable on large graphs.

\begin{acks}
This work was supported in part by National Science Foundation grants
IIS-1908166 and IIS-1908198.
\end{acks}

%
\bibliographystyle{ACM-Reference-Format}
\bibliography{deep}


\begin{thebibliography}{37}


\ifx \showCODEN    \undefined \def \showCODEN     #1{\unskip}     \fi
\ifx \showDOI      \undefined \def \showDOI       #1{#1}\fi
\ifx \showISBNx    \undefined \def \showISBNx     #1{\unskip}     \fi
\ifx \showISBNxiii \undefined \def \showISBNxiii  #1{\unskip}     \fi
\ifx \showISSN     \undefined \def \showISSN      #1{\unskip}     \fi
\ifx \showLCCN     \undefined \def \showLCCN      #1{\unskip}     \fi
\ifx \shownote     \undefined \def \shownote      #1{#1}          \fi
\ifx \showarticletitle \undefined \def \showarticletitle #1{#1}   \fi
\ifx \showURL      \undefined \def \showURL       {\relax}        \fi
\providecommand\bibfield[2]{#2}
\providecommand\bibinfo[2]{#2}
\providecommand\natexlab[1]{#1}
\providecommand\showeprint[2][]{arXiv:#2}

\bibitem[\protect\citeauthoryear{Abadi, Barham, Chen, Chen, Davis, Dean, Devin,
  Ghemawat, Irving, Isard, et~al\mbox{.}}{Abadi et~al\mbox{.}}{2016}]%
        {abadi2016tensorflow}
\bibfield{author}{\bibinfo{person}{Mart{\'\i}n Abadi}, \bibinfo{person}{Paul
  Barham}, \bibinfo{person}{Jianmin Chen}, \bibinfo{person}{Zhifeng Chen},
  \bibinfo{person}{Andy Davis}, \bibinfo{person}{Jeffrey Dean},
  \bibinfo{person}{Matthieu Devin}, \bibinfo{person}{Sanjay Ghemawat},
  \bibinfo{person}{Geoffrey Irving}, \bibinfo{person}{Michael Isard},
  {et~al\mbox{.}}} \bibinfo{year}{2016}\natexlab{}.
\newblock \showarticletitle{Tensorflow: a system for large-scale machine
  learning}. In \bibinfo{booktitle}{\emph{OSDI}}, Vol.~\bibinfo{volume}{16}.
  \bibinfo{pages}{265--283}.
\newblock


\bibitem[\protect\citeauthoryear{Borgwardt, Ong, Sch{\"o}nauer, Vishwanathan,
  Smola, and Kriegel}{Borgwardt et~al\mbox{.}}{2005}]%
        {borgwardt2005protein}
\bibfield{author}{\bibinfo{person}{Karsten~M Borgwardt},
  \bibinfo{person}{Cheng~Soon Ong}, \bibinfo{person}{Stefan Sch{\"o}nauer},
  \bibinfo{person}{SVN Vishwanathan}, \bibinfo{person}{Alex~J Smola}, {and}
  \bibinfo{person}{Hans-Peter Kriegel}.} \bibinfo{year}{2005}\natexlab{}.
\newblock \showarticletitle{Protein function prediction via graph kernels}.
\newblock \bibinfo{journal}{\emph{Bioinformatics}} \bibinfo{volume}{21},
  \bibinfo{number}{suppl\_1} (\bibinfo{year}{2005}), \bibinfo{pages}{i47--i56}.
\newblock


\bibitem[\protect\citeauthoryear{Chang and Lin}{Chang and Lin}{2011}]%
        {chang2011libsvm}
\bibfield{author}{\bibinfo{person}{Chih-Chung Chang} {and}
  \bibinfo{person}{Chih-Jen Lin}.} \bibinfo{year}{2011}\natexlab{}.
\newblock \showarticletitle{LIBSVM: a library for support vector machines}.
\newblock \bibinfo{journal}{\emph{ACM Transactions on Intelligent Systems and
  Technology}} \bibinfo{volume}{2}, \bibinfo{number}{3} (\bibinfo{year}{2011}),
  \bibinfo{pages}{27}.
\newblock


\bibitem[\protect\citeauthoryear{Defferrard, Bresson, and
  Vandergheynst}{Defferrard et~al\mbox{.}}{2016}]%
        {defferrard2016convolutional}
\bibfield{author}{\bibinfo{person}{Micha{\"e}l Defferrard},
  \bibinfo{person}{Xavier Bresson}, {and} \bibinfo{person}{Pierre
  Vandergheynst}.} \bibinfo{year}{2016}\natexlab{}.
\newblock \showarticletitle{Convolutional neural networks on graphs with fast
  localized spectral filtering}. In \bibinfo{booktitle}{\emph{Advances in
  Neural Information Processing Systems}}. \bibinfo{pages}{3844--3852}.
\newblock


\bibitem[\protect\citeauthoryear{Devlin, Chang, Lee, and Toutanova}{Devlin
  et~al\mbox{.}}{2018}]%
        {devlin2018bert}
\bibfield{author}{\bibinfo{person}{Jacob Devlin}, \bibinfo{person}{Ming-Wei
  Chang}, \bibinfo{person}{Kenton Lee}, {and} \bibinfo{person}{Kristina
  Toutanova}.} \bibinfo{year}{2018}\natexlab{}.
\newblock \showarticletitle{Bert: Pre-training of deep bidirectional
  transformers for language understanding}.
\newblock \bibinfo{journal}{\emph{arXiv preprint arXiv:1810.04805}}
  (\bibinfo{year}{2018}).
\newblock


\bibitem[\protect\citeauthoryear{Dobson and Doig}{Dobson and Doig}{2003}]%
        {dobson2003distinguishing}
\bibfield{author}{\bibinfo{person}{Paul~D Dobson} {and}
  \bibinfo{person}{Andrew~J Doig}.} \bibinfo{year}{2003}\natexlab{}.
\newblock \showarticletitle{Distinguishing enzyme structures from non-enzymes
  without alignments}.
\newblock \bibinfo{journal}{\emph{Journal of Molecular Biology}}
  \bibinfo{volume}{330}, \bibinfo{number}{4} (\bibinfo{year}{2003}),
  \bibinfo{pages}{771--783}.
\newblock


\bibitem[\protect\citeauthoryear{Gao, Wang, and Ji}{Gao et~al\mbox{.}}{2018}]%
        {gao2018large}
\bibfield{author}{\bibinfo{person}{Hongyang Gao}, \bibinfo{person}{Zhengyang
  Wang}, {and} \bibinfo{person}{Shuiwang Ji}.} \bibinfo{year}{2018}\natexlab{}.
\newblock \showarticletitle{Large-scale learnable graph convolutional
  networks}. In \bibinfo{booktitle}{\emph{Proceedings of the 24th ACM SIGKDD
  International Conference on Knowledge Discovery \& Data Mining}}. ACM,
  \bibinfo{pages}{1416--1424}.
\newblock


\bibitem[\protect\citeauthoryear{Glorot and Bengio}{Glorot and Bengio}{2010}]%
        {glorot2010understanding}
\bibfield{author}{\bibinfo{person}{Xavier Glorot} {and} \bibinfo{person}{Yoshua
  Bengio}.} \bibinfo{year}{2010}\natexlab{}.
\newblock \showarticletitle{Understanding the difficulty of training deep
  feedforward neural networks}. In \bibinfo{booktitle}{\emph{Proceedings of the
  Thirteenth International Conference on Artificial Intelligence and
  Statistics}}. \bibinfo{pages}{249--256}.
\newblock


\bibitem[\protect\citeauthoryear{Gregor, Danihelka, Graves, Rezende, and
  Wierstra}{Gregor et~al\mbox{.}}{2015}]%
        {gregor2015draw}
\bibfield{author}{\bibinfo{person}{Karol Gregor}, \bibinfo{person}{Ivo
  Danihelka}, \bibinfo{person}{Alex Graves}, \bibinfo{person}{Danilo Rezende},
  {and} \bibinfo{person}{Daan Wierstra}.} \bibinfo{year}{2015}\natexlab{}.
\newblock \showarticletitle{DRAW: A recurrent neural network for image
  generation}. In \bibinfo{booktitle}{\emph{International Conference on Machine
  Learning}}. \bibinfo{pages}{1462--1471}.
\newblock


\bibitem[\protect\citeauthoryear{Hamilton, Ying, and Leskovec}{Hamilton
  et~al\mbox{.}}{2017}]%
        {hamilton2017inductive}
\bibfield{author}{\bibinfo{person}{Will Hamilton}, \bibinfo{person}{Zhitao
  Ying}, {and} \bibinfo{person}{Jure Leskovec}.}
  \bibinfo{year}{2017}\natexlab{}.
\newblock \showarticletitle{Inductive representation learning on large graphs}.
  In \bibinfo{booktitle}{\emph{Advances in Neural Information Processing
  Systems}}. \bibinfo{pages}{1024--1034}.
\newblock


\bibitem[\protect\citeauthoryear{Hochreiter and Schmidhuber}{Hochreiter and
  Schmidhuber}{1997}]%
        {hochreiter1997long}
\bibfield{author}{\bibinfo{person}{Sepp Hochreiter} {and}
  \bibinfo{person}{J{\"u}rgen Schmidhuber}.} \bibinfo{year}{1997}\natexlab{}.
\newblock \showarticletitle{Long short-term memory}.
\newblock \bibinfo{journal}{\emph{Neural Computation}} \bibinfo{volume}{9},
  \bibinfo{number}{8} (\bibinfo{year}{1997}), \bibinfo{pages}{1735--1780}.
\newblock


\bibitem[\protect\citeauthoryear{Jaderberg, Simonyan, Zisserman,
  et~al\mbox{.}}{Jaderberg et~al\mbox{.}}{2015}]%
        {jaderberg2015spatial}
\bibfield{author}{\bibinfo{person}{Max Jaderberg}, \bibinfo{person}{Karen
  Simonyan}, \bibinfo{person}{Andrew Zisserman}, {et~al\mbox{.}}}
  \bibinfo{year}{2015}\natexlab{}.
\newblock \showarticletitle{Spatial transformer networks}. In
  \bibinfo{booktitle}{\emph{Advances in Neural Information Processing
  Systems}}. \bibinfo{pages}{2017--2025}.
\newblock


\bibitem[\protect\citeauthoryear{Juefei-Xu, Verma, Goel, Cherodian, and
  Savvides}{Juefei-Xu et~al\mbox{.}}{2016}]%
        {juefei2016deepgender}
\bibfield{author}{\bibinfo{person}{Felix Juefei-Xu}, \bibinfo{person}{Eshan
  Verma}, \bibinfo{person}{Parag Goel}, \bibinfo{person}{Anisha Cherodian},
  {and} \bibinfo{person}{Marios Savvides}.} \bibinfo{year}{2016}\natexlab{}.
\newblock \showarticletitle{Deepgender: Occlusion and low resolution robust
  facial gender classification via progressively trained convolutional neural
  networks with attention}. In \bibinfo{booktitle}{\emph{Proceedings of the
  IEEE Conference on Computer Vision and Pattern Recognition Workshops}}.
  \bibinfo{pages}{68--77}.
\newblock


\bibitem[\protect\citeauthoryear{Kingma and Ba}{Kingma and Ba}{2015}]%
        {kingma2014adam}
\bibfield{author}{\bibinfo{person}{Diederik Kingma} {and}
  \bibinfo{person}{Jimmy Ba}.} \bibinfo{year}{2015}\natexlab{}.
\newblock \showarticletitle{Adam: A method for stochastic optimization}.
\newblock \bibinfo{journal}{\emph{The International Conference on Learning
  Representations}} (\bibinfo{year}{2015}).
\newblock


\bibitem[\protect\citeauthoryear{Kipf and Welling}{Kipf and Welling}{2017}]%
        {kipf2016semi}
\bibfield{author}{\bibinfo{person}{Thomas~N Kipf} {and} \bibinfo{person}{Max
  Welling}.} \bibinfo{year}{2017}\natexlab{}.
\newblock \showarticletitle{Semi-supervised classification with graph
  convolutional networks}.
\newblock \bibinfo{journal}{\emph{International Conference on Learning
  Representations}} (\bibinfo{year}{2017}).
\newblock


\bibitem[\protect\citeauthoryear{LeCun, Bottou, Orr, and M{\"u}ller}{LeCun
  et~al\mbox{.}}{2012}]%
        {lecun2012efficient}
\bibfield{author}{\bibinfo{person}{Yann LeCun}, \bibinfo{person}{L{\'e}on
  Bottou}, \bibinfo{person}{Genevieve~B Orr}, {and}
  \bibinfo{person}{Klaus-Robert M{\"u}ller}.} \bibinfo{year}{2012}\natexlab{}.
\newblock \showarticletitle{Efficient backprop}.
\newblock In \bibinfo{booktitle}{\emph{Neural networks: Tricks of the trade}}.
  \bibinfo{publisher}{Springer}, \bibinfo{pages}{9--48}.
\newblock


\bibitem[\protect\citeauthoryear{Li, He, Zhang, Chang, Dong, and Lin}{Li
  et~al\mbox{.}}{2018}]%
        {li2018non}
\bibfield{author}{\bibinfo{person}{Guanbin Li}, \bibinfo{person}{Xiang He},
  \bibinfo{person}{Wei Zhang}, \bibinfo{person}{Huiyou Chang},
  \bibinfo{person}{Le Dong}, {and} \bibinfo{person}{Liang Lin}.}
  \bibinfo{year}{2018}\natexlab{}.
\newblock \showarticletitle{Non-locally enhanced encoder-decoder network for
  single image de-raining}. In \bibinfo{booktitle}{\emph{2018 ACM Multimedia
  Conference on Multimedia Conference}}. ACM, \bibinfo{pages}{1056--1064}.
\newblock


\bibitem[\protect\citeauthoryear{Ling and Rush}{Ling and Rush}{2017}]%
        {ling2017coarse}
\bibfield{author}{\bibinfo{person}{Jeffrey Ling} {and}
  \bibinfo{person}{Alexander Rush}.} \bibinfo{year}{2017}\natexlab{}.
\newblock \showarticletitle{Coarse-to-fine attention models for document
  summarization}. In \bibinfo{booktitle}{\emph{Proceedings of the Workshop on
  New Frontiers in Summarization}}. \bibinfo{pages}{33--42}.
\newblock


\bibitem[\protect\citeauthoryear{Luong, Pham, and Manning}{Luong
  et~al\mbox{.}}{2015}]%
        {luong2015effective}
\bibfield{author}{\bibinfo{person}{Thang Luong}, \bibinfo{person}{Hieu Pham},
  {and} \bibinfo{person}{Christopher~D Manning}.}
  \bibinfo{year}{2015}\natexlab{}.
\newblock \showarticletitle{Effective approaches to attention-based neural
  machine translation}. In \bibinfo{booktitle}{\emph{Proceedings of the 2015
  Conference on Empirical Methods in Natural Language Processing}}.
  \bibinfo{pages}{1412--1421}.
\newblock


\bibitem[\protect\citeauthoryear{Malinowski, Doersch, Santoro, and
  Battaglia}{Malinowski et~al\mbox{.}}{2018}]%
        {malinowski2018learning}
\bibfield{author}{\bibinfo{person}{Mateusz Malinowski}, \bibinfo{person}{Carl
  Doersch}, \bibinfo{person}{Adam Santoro}, {and} \bibinfo{person}{Peter
  Battaglia}.} \bibinfo{year}{2018}\natexlab{}.
\newblock \showarticletitle{Learning visual question answering by bootstrapping
  hard attention}. In \bibinfo{booktitle}{\emph{European Conference on Computer
  Vision}}. Springer, \bibinfo{pages}{3--20}.
\newblock


\bibitem[\protect\citeauthoryear{Niepert, Ahmed, and Kutzkov}{Niepert
  et~al\mbox{.}}{2016}]%
        {niepert2016learning}
\bibfield{author}{\bibinfo{person}{Mathias Niepert}, \bibinfo{person}{Mohamed
  Ahmed}, {and} \bibinfo{person}{Konstantin Kutzkov}.}
  \bibinfo{year}{2016}\natexlab{}.
\newblock \showarticletitle{Learning convolutional neural networks for graphs}.
  In \bibinfo{booktitle}{\emph{International Conference on Machine Learning}}.
  \bibinfo{pages}{2014--2023}.
\newblock


\bibitem[\protect\citeauthoryear{Perozzi, Al-Rfou, and Skiena}{Perozzi
  et~al\mbox{.}}{2014}]%
        {perozzi2014deepwalk}
\bibfield{author}{\bibinfo{person}{Bryan Perozzi}, \bibinfo{person}{Rami
  Al-Rfou}, {and} \bibinfo{person}{Steven Skiena}.}
  \bibinfo{year}{2014}\natexlab{}.
\newblock \showarticletitle{Deepwalk: Online learning of social
  representations}. In \bibinfo{booktitle}{\emph{Proceedings of the 20th ACM
  SIGKDD International Conference on Knowledge Discovery and Data Mining}}.
  \bibinfo{pages}{701--710}.
\newblock


\bibitem[\protect\citeauthoryear{Rao, Lu, and Zhou}{Rao et~al\mbox{.}}{2017}]%
        {rao2017attention}
\bibfield{author}{\bibinfo{person}{Yongming Rao}, \bibinfo{person}{Jiwen Lu},
  {and} \bibinfo{person}{Jie Zhou}.} \bibinfo{year}{2017}\natexlab{}.
\newblock \showarticletitle{Attention-aware deep reinforcement learning for
  video face recognition}. In \bibinfo{booktitle}{\emph{Proceedings of the IEEE
  Conference on Computer Vision and Pattern Recognition}}.
  \bibinfo{pages}{3931--3940}.
\newblock


\bibitem[\protect\citeauthoryear{Sen, Namata, Bilgic, Getoor, Galligher, and
  Eliassi-Rad}{Sen et~al\mbox{.}}{2008}]%
        {sen2008collective}
\bibfield{author}{\bibinfo{person}{Prithviraj Sen}, \bibinfo{person}{Galileo
  Namata}, \bibinfo{person}{Mustafa Bilgic}, \bibinfo{person}{Lise Getoor},
  \bibinfo{person}{Brian Galligher}, {and} \bibinfo{person}{Tina Eliassi-Rad}.}
  \bibinfo{year}{2008}\natexlab{}.
\newblock \showarticletitle{Collective classification in network data}.
\newblock \bibinfo{journal}{\emph{AI Magazine}} \bibinfo{volume}{29},
  \bibinfo{number}{3} (\bibinfo{year}{2008}), \bibinfo{pages}{93}.
\newblock


\bibitem[\protect\citeauthoryear{Shankar, Garg, and Sarawagi}{Shankar
  et~al\mbox{.}}{2018}]%
        {shankar2018surprisingly}
\bibfield{author}{\bibinfo{person}{Shiv Shankar}, \bibinfo{person}{Siddhant
  Garg}, {and} \bibinfo{person}{Sunita Sarawagi}.}
  \bibinfo{year}{2018}\natexlab{}.
\newblock \showarticletitle{Surprisingly easy hard-attention for sequence to
  sequence learning}. In \bibinfo{booktitle}{\emph{Proceedings of the 2018
  Conference on Empirical Methods in Natural Language Processing}}.
  \bibinfo{pages}{640--645}.
\newblock


\bibitem[\protect\citeauthoryear{Srivastava, Hinton, Krizhevsky, Sutskever, and
  Salakhutdinov}{Srivastava et~al\mbox{.}}{2014}]%
        {srivastava2014dropout}
\bibfield{author}{\bibinfo{person}{Nitish Srivastava},
  \bibinfo{person}{Geoffrey Hinton}, \bibinfo{person}{Alex Krizhevsky},
  \bibinfo{person}{Ilya Sutskever}, {and} \bibinfo{person}{Ruslan
  Salakhutdinov}.} \bibinfo{year}{2014}\natexlab{}.
\newblock \showarticletitle{Dropout: A simple way to prevent neural networks
  from overfitting}.
\newblock \bibinfo{journal}{\emph{Journal of Machine Learning Research}}
  \bibinfo{volume}{15}, \bibinfo{number}{1} (\bibinfo{year}{2014}),
  \bibinfo{pages}{1929--1958}.
\newblock


\bibitem[\protect\citeauthoryear{Vaswani, Shazeer, Parmar, Uszkoreit, Jones,
  Gomez, Kaiser, and Polosukhin}{Vaswani et~al\mbox{.}}{2017}]%
        {vaswani2017attention}
\bibfield{author}{\bibinfo{person}{Ashish Vaswani}, \bibinfo{person}{Noam
  Shazeer}, \bibinfo{person}{Niki Parmar}, \bibinfo{person}{Jakob Uszkoreit},
  \bibinfo{person}{Llion Jones}, \bibinfo{person}{Aidan~N Gomez},
  \bibinfo{person}{{\L}ukasz Kaiser}, {and} \bibinfo{person}{Illia
  Polosukhin}.} \bibinfo{year}{2017}\natexlab{}.
\newblock \showarticletitle{Attention is all you need}. In
  \bibinfo{booktitle}{\emph{Advances in Neural Information Processing
  Systems}}. \bibinfo{pages}{6000--6010}.
\newblock


\bibitem[\protect\citeauthoryear{Veli{\v{c}}kovi{\'c}, Cucurull, Casanova,
  Romero, Li{\`o}, and Bengio}{Veli{\v{c}}kovi{\'c} et~al\mbox{.}}{2017}]%
        {velivckovic2017graph}
\bibfield{author}{\bibinfo{person}{Petar Veli{\v{c}}kovi{\'c}},
  \bibinfo{person}{Guillem Cucurull}, \bibinfo{person}{Arantxa Casanova},
  \bibinfo{person}{Adriana Romero}, \bibinfo{person}{Pietro Li{\`o}}, {and}
  \bibinfo{person}{Yoshua Bengio}.} \bibinfo{year}{2017}\natexlab{}.
\newblock \showarticletitle{Graph attention networks}. In
  \bibinfo{booktitle}{\emph{International Conference on Learning
  Representations}}.
\newblock


\bibitem[\protect\citeauthoryear{Vinyals, Bengio, and Kudlur}{Vinyals
  et~al\mbox{.}}{2016}]%
        {vinyals2015order}
\bibfield{author}{\bibinfo{person}{Oriol Vinyals}, \bibinfo{person}{Samy
  Bengio}, {and} \bibinfo{person}{Manjunath Kudlur}.}
  \bibinfo{year}{2016}\natexlab{}.
\newblock \showarticletitle{Order matters: Sequence to sequence for sets}.
\newblock \bibinfo{journal}{\emph{International Conference on Learning
  Representations}} (\bibinfo{year}{2016}).
\newblock


\bibitem[\protect\citeauthoryear{Wang, Girshick, Gupta, and He}{Wang
  et~al\mbox{.}}{2018}]%
        {wang2018non}
\bibfield{author}{\bibinfo{person}{Xiaolong Wang}, \bibinfo{person}{Ross
  Girshick}, \bibinfo{person}{Abhinav Gupta}, {and} \bibinfo{person}{Kaiming
  He}.} \bibinfo{year}{2018}\natexlab{}.
\newblock \showarticletitle{Non-local neural networks}. In
  \bibinfo{booktitle}{\emph{The IEEE Conference on Computer Vision and Pattern
  Recognition}}, Vol.~\bibinfo{volume}{1}. \bibinfo{pages}{4}.
\newblock


\bibitem[\protect\citeauthoryear{Xu, Ba, Kiros, Cho, Courville, Salakhudinov,
  Zemel, and Bengio}{Xu et~al\mbox{.}}{2015}]%
        {xu2015show}
\bibfield{author}{\bibinfo{person}{Kelvin Xu}, \bibinfo{person}{Jimmy Ba},
  \bibinfo{person}{Ryan Kiros}, \bibinfo{person}{Kyunghyun Cho},
  \bibinfo{person}{Aaron Courville}, \bibinfo{person}{Ruslan Salakhudinov},
  \bibinfo{person}{Rich Zemel}, {and} \bibinfo{person}{Yoshua Bengio}.}
  \bibinfo{year}{2015}\natexlab{}.
\newblock \showarticletitle{Show, attend and tell: Neural image caption
  generation with visual attention}. In \bibinfo{booktitle}{\emph{International
  Conference on Machine Learning}}. \bibinfo{pages}{2048--2057}.
\newblock


\bibitem[\protect\citeauthoryear{Yanardag and Vishwanathan}{Yanardag and
  Vishwanathan}{2015}]%
        {yanardag2015structural}
\bibfield{author}{\bibinfo{person}{Pinar Yanardag} {and} \bibinfo{person}{SVN
  Vishwanathan}.} \bibinfo{year}{2015}\natexlab{}.
\newblock \showarticletitle{A structural smoothing framework for robust graph
  comparison}. In \bibinfo{booktitle}{\emph{Advances in Neural Information
  Processing Systems}}. \bibinfo{pages}{2134--2142}.
\newblock


\bibitem[\protect\citeauthoryear{Yang, Cohen, and Salakhudinov}{Yang
  et~al\mbox{.}}{2016}]%
        {yang2016revisiting}
\bibfield{author}{\bibinfo{person}{Zhilin Yang}, \bibinfo{person}{William
  Cohen}, {and} \bibinfo{person}{Ruslan Salakhudinov}.}
  \bibinfo{year}{2016}\natexlab{}.
\newblock \showarticletitle{Revisiting semi-supervised learning with graph
  embeddings}. In \bibinfo{booktitle}{\emph{International Conference on Machine
  Learning}}. \bibinfo{pages}{40--48}.
\newblock


\bibitem[\protect\citeauthoryear{Ying, You, Morris, Ren, Hamilton, and
  Leskovec}{Ying et~al\mbox{.}}{2018}]%
        {ying2018hierarchical}
\bibfield{author}{\bibinfo{person}{Zhitao Ying}, \bibinfo{person}{Jiaxuan You},
  \bibinfo{person}{Christopher Morris}, \bibinfo{person}{Xiang Ren},
  \bibinfo{person}{Will Hamilton}, {and} \bibinfo{person}{Jure Leskovec}.}
  \bibinfo{year}{2018}\natexlab{}.
\newblock \showarticletitle{Hierarchical graph representation learning with
  differentiable pooling}. In \bibinfo{booktitle}{\emph{Advances in Neural
  Information Processing Systems}}. \bibinfo{pages}{4800--4810}.
\newblock


\bibitem[\protect\citeauthoryear{Yu, Yin, and Zhu}{Yu et~al\mbox{.}}{2019}]%
        {yu2019st}
\bibfield{author}{\bibinfo{person}{Bing Yu}, \bibinfo{person}{Haoteng Yin},
  {and} \bibinfo{person}{Zhanxing Zhu}.} \bibinfo{year}{2019}\natexlab{}.
\newblock \showarticletitle{ST-UNet: A spatio-temporal {U}-network for
  graph-structured time series modeling}.
\newblock \bibinfo{journal}{\emph{arXiv preprint arXiv:1903.05631}}
  (\bibinfo{year}{2019}).
\newblock


\bibitem[\protect\citeauthoryear{Zhang, Cui, Neumann, and Chen}{Zhang
  et~al\mbox{.}}{2018}]%
        {zhang2018end}
\bibfield{author}{\bibinfo{person}{Muhan Zhang}, \bibinfo{person}{Zhicheng
  Cui}, \bibinfo{person}{Marion Neumann}, {and} \bibinfo{person}{Yixin Chen}.}
  \bibinfo{year}{2018}\natexlab{}.
\newblock \showarticletitle{An end-to-end deep learning architecture for graph
  classification}. In \bibinfo{booktitle}{\emph{Proceedings of AAAI Conference
  on Artificial Inteligence}}.
\newblock


\bibitem[\protect\citeauthoryear{Zhao, Zhang, Liu, Shi, Change~Loy, Lin, and
  Jia}{Zhao et~al\mbox{.}}{2018}]%
        {zhao2018psanet}
\bibfield{author}{\bibinfo{person}{Hengshuang Zhao}, \bibinfo{person}{Yi
  Zhang}, \bibinfo{person}{Shu Liu}, \bibinfo{person}{Jianping Shi},
  \bibinfo{person}{Chen Change~Loy}, \bibinfo{person}{Dahua Lin}, {and}
  \bibinfo{person}{Jiaya Jia}.} \bibinfo{year}{2018}\natexlab{}.
\newblock \showarticletitle{Psanet: Point-wise spatial attention network for
  scene parsing}. In \bibinfo{booktitle}{\emph{Proceedings of the European
  Conference on Computer Vision}}. \bibinfo{pages}{267--283}.
\newblock


\end{thebibliography}

\end{document}